\documentclass{article}

 \usepackage[main, final]{neurips_2026}

\usepackage{graphicx}
\usepackage{subfigure}
\usepackage{algorithm}
\usepackage{algpseudocode}
\usepackage{amsmath}
\usepackage{hyperref}
\usepackage{pifont}
\usepackage{wrapfig}
\usepackage{lipsum}
\usepackage{caption}
\usepackage{float}

\usepackage[utf8]{inputenc} 
\usepackage[T1]{fontenc}    
\usepackage{hyperref}       
\usepackage{url}            
\usepackage{booktabs}       
\usepackage{amsfonts}       
\usepackage{nicefrac}       
\usepackage{microtype}      
\usepackage{xcolor}         

\usepackage{amsmath,amsfonts,bm}

\def\eqref#1{equation~\ref{#1}}

\def\1{\bm{1}}

\DeclareMathAlphabet{\mathsfit}{\encodingdefault}{\sfdefault}{m}{sl}
\SetMathAlphabet{\mathsfit}{bold}{\encodingdefault}{\sfdefault}{bx}{n}

\title{CompactSplat: Spatially Adaptive Gaussian Distribution for Feedforward Scene Reconstruction}

\author{Bing He 
\textsuperscript{1}
\& Jingnan Gao
\textsuperscript{1}
\& Yunuo Chen
\textsuperscript{1}
\\
\textbf{\& Zhengxue Cheng
\textsuperscript{1}
\& Rong Xie
\textsuperscript{1}
\& Li Song
\textsuperscript{1}
\& Wenjun Zhang 
\textsuperscript{1}}
\\
\textsuperscript{1} Shanghai Jiao Tong University
\\
\texttt{\{sandwich\_theorem\}@sjtu.edu.cn} \\
}

\begin{document}

\newcounter{mypara}
\maketitle

\begin{center}
    \vspace{-20pt}
    \centering
    \captionsetup{type=figure}
    \includegraphics[width=\textwidth]{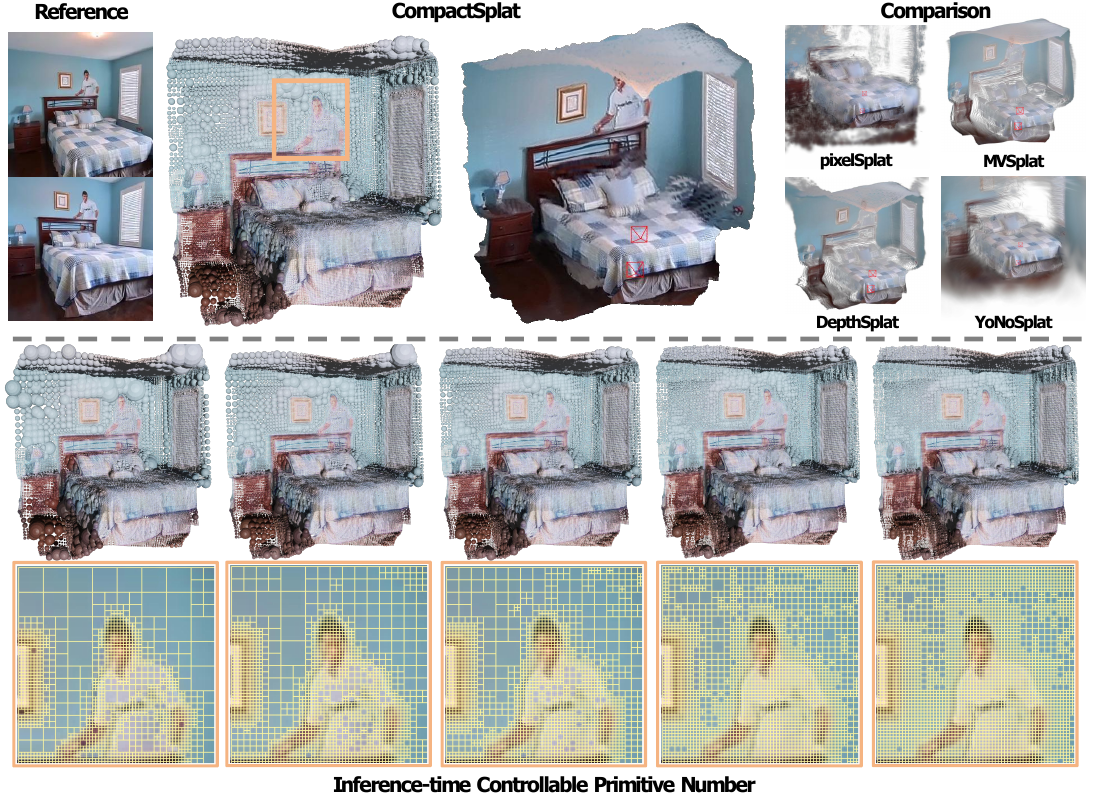}
    \captionof{figure}{
    \textbf{CompactSplat} is a highly efficient feed-forward network that predicts 3D scene representations from sparse input images. While previous methods typically allocate one or more Gaussians per pixel—leading to excessive primitives and spatially redundant representations, CompactSplat adaptively allocates primitives based on regional complexity. It concentrates dense Gaussians in texturally rich areas while deploying expanded, scale-modulated primitives in flat regions. This paradigm enables high-fidelity reconstruction with substantially fewer primitives. Crucially, it inherently supports dynamic sparsity control at inference time, allowing seamless trade-offs between visual fidelity and memory footprint across diverse deployment constraints.
    }
    \label{teaser}
\end{center}%

\begin{abstract}
\refstepcounter{mypara} \label{para:abstract}
Reconstructing 3D scenes from sparse views without per-scene optimization remains highly challenging, especially for recovering accurate geometry and fine textures.
While recent feedforward approaches leverage generalizable 3D Gaussian Splatting (3DGS) for scene generation, they typically assign one or multiple Gaussians to each pixel. Such uniform allocation not only produces highly redundant representations by wasting primitives in homogeneous areas, but also fails to exploit the inherent geometric priors of planar surfaces, often resulting in sub-optimal structural modeling. To address these limitations, we present \textbf{CompactSplat}, a novel feedforward framework that enables a content-aware allocation of 3D Gaussians. By integrating texture-guided spatial partitioning with hierarchical sampling, CompactSplat adaptively distributes primitives — concentrating dense Gaussians in complex regions while tiling flat areas with expanded, scale-modulated primitives. Moreover, our framework inherently supports dynamic sparsity control at inference time, enabling seamless trade-offs between rendering fidelity and memory footprint without requiring network retraining. Extensive experiments on RealEstate10K, DL3DV, and ScanNet demonstrate that CompactSplat consistently outperforms prior methods on both standard metrics and high-resolution rendering consistency, achieving high-fidelity reconstructions with significantly fewer primitives.

\end{abstract}
\vspace{-15pt}    
\section{Introduction}
\vspace{-5pt}
Recovering geometrically accurate and visually consistent 3D scenes from sparse views is a fundamental problem for immersive VR, gaming, and digital content creation. The recent success of 3D Gaussian Splatting (3DGS)~\cite{kerbl20233d} has significantly advanced this field by representing scenes as collections of 3D Gaussians, enabling real-time rendering with high visual fidelity.

Existing methods for constructing such representations generally follow two paradigms. Traditional approaches rely on dense multi-view captures and COLMAP-based~\cite{schoenberger2016sfm} point cloud initialization, followed by computationally expensive per-scene optimization. To avoid this bottleneck, feed-forward approaches have recently gained increasing attention. By leveraging pretrained models, they directly predict 3D Gaussians from sparse inputs in a single forward pass, eliminating per-scene optimization and enabling fast, scalable 3D reconstruction.

Despite their rapid inference, many existing feed-forward methods remain inefficient in their primitive allocation. Natural scenes exhibit substantial spatial variation in texture and geometric complexity, yet prior methods typically adopt a rigid pixel-wise prediction paradigm, assigning one~\cite{chen2024mvsplat,xu2024depthsplat,he2026surfsplat} or multiple~\cite{charatan2024pixelsplat,tang2024hisplat} Gaussians to every pixel. This uniform allocation forces the network to populate large, textureless regions with thousands of primitives. Consequently, the generated 3D representations can suffer from substantial spatial redundancy. This redundancy increases representation storage costs and motivates more compact 3D priors for downstream tasks.

To address this issue, we introduce \textbf{CompactSplat}, a novel feed-forward framework for high-fidelity 3D reconstruction with content-aware Gaussian allocation. Instead of retaining a primitive at every pixel, CompactSplat allocates Gaussians according to image texture, using sparse primitives in homogeneous regions while concentrating denser predictions in areas with rich texture. 
To achieve this adaptive allocation, we decouple the primitive generation process into two complementary components. First, a \textbf{Texture-Aware Partitioner (TAP)} analyzes the input views to estimate regional texture complexity and partition each image with a quadtree. Second, a \textbf{Hierarchical Pyramid Sampler (HPS)} uses this guidance to sample explicit Gaussian attributes from a pooled multi-resolution attribute pyramid. Together, these components enable CompactSplat to adapt its primitive density to the input texture, reducing spatial redundancy while retaining strong reconstruction quality.

Crucially, CompactSplat provides a \textbf{unified paradigm for controllable inference}. Owing to the scale-aware synergy between TAP and HPS, our framework naturally supports dynamic sparsity control at inference time. By modulating a single partition threshold, the model can select different trade-offs between rendering fidelity and primitive count---\textbf{without requiring any network retraining or auxiliary modules.} This intrinsic adaptability provides an on-the-fly mechanism to select different representation sizes using a single set of pretrained weights.

Ultimately, CompactSplat establishes a more compact paradigm for 3D representation. By replacing rigid pixel-wise allocation with texture-driven primitive allocation, our approach adapts to spatial variations in image content. In our experiments, it delivers high-fidelity reconstructions with fewer primitives. This allocation transforms dense, redundant outputs into compact 3D representations with a tunable primitive count.

In summary, our contributions are as follows:
\begin{itemize}
    \item We propose \textbf{CompactSplat}, a feed-forward framework for compact sparse-view 3D reconstruction. By combining a Texture-Aware Partitioner (TAP) with a Hierarchical Pyramid Sampler (HPS), our method adaptively allocates 3D Gaussians based on image texture, significantly reducing spatial redundancy.
    
    \item  We introduce a \textbf{controllable inference paradigm} for feed-forward 3D Gaussian Splatting. With a single partition threshold, CompactSplat supports dynamic, on-the-fly sparsity control. This capability enables seamless trade-offs between visual fidelity and primitive count using a single set of network weights, completely eliminating the need for sparsity-specific retraining.
    
    \item Extensive experiments on RealEstate10K and ACID demonstrate that CompactSplat achieves \textbf{competitive quality with fewer primitives}. It retains strong performance on both standard metrics and high-resolution rendering consistency, while cross-dataset evaluation on DTU, DL3DV, and ScanNet reveals quality--compactness trade-offs.
\end{itemize}

\vspace{-15pt}
\section{Related Works}
\vspace{-5pt}

\subsection{3D Gaussian Splatting}
\vspace{-5pt}
Since the advent of Neural Radiance Fields (NeRF)~\cite{mildenhall2021nerf}, novel view synthesis has advanced rapidly, evolving from implicit MLPs to grid-based~\cite{muller2022instant,fridovich2022plenoxels} and anti-aliased~\cite{barron2021mip,barron2022mip} representations. To further accelerate rendering, recent paradigms shifted towards explicit point-based architectures. Notably, 3D Gaussian Splatting (3DGS)~\cite{kerbl20233d} revolutionized the field by modeling scenes with semi-transparent, anisotropic ellipsoids, enabling photorealistic, real-time rendering via differentiable rasterization. Consequently, a proliferation of extensions has emerged to refine 3DGS. These include optimizations for training stability~\cite{cheng2024gaussianpro,zhang2024fregs,radl2024stopthepop,diolatzis2024n}, anti-aliasing techniques~\cite{yan2024multi,yu2024mip,song2024sa,liang2024analytic}, and adaptations for dynamic scenes~\cite{luiten2023dynamic,wu20234d,wan2024superpoint,huang2023sc,he2024s4d}. Additionally, several works focus on compressing these representations to reduce storage and accelerate inference~\cite{girish2024eagles,navaneet2024compgs,niedermayr2024compressed,lee2024compact,fan2024lightgaussian,chen2024hac}.
\vspace{-10pt}
\subsection{Generalizable 3D Reconstruction}
\vspace{-5pt}
To bypass the computationally expensive per-scene optimization of traditional 3DGS, recent literature has explored feedforward models that directly predict Gaussian attributes from sparse inputs. Splatter Image~\cite{szymanowicz2024splatter} pioneered the translation of 2D images into Gaussian parameter maps. Subsequent frameworks integrated geometric priors to enhance spatial understanding; for instance, pixelSplat~\cite{charatan2024pixelsplat} utilized epipolar geometry, while MVSplat~\cite{chen2024mvsplat} and DepthSplat~\cite{xu2024depthsplat} leveraged cost volumes and cross-task depth interactions. Further advancements addressed specific challenges, such as expanding synthesis ranges~\cite{wang2024freesplat}, improving geometric surfaces via 2DGS~\cite{he2026surfsplat}, and handling unposed inputs~\cite{kang2025selfsplat,jiang2025anysplat}. VGGT~\cite{wang2025vggt} and MV-DUSt3R~\cite{tang2025mv} likewise reconstruct uncalibrated views, while CompactSplat assumes known camera intrinsics and extrinsics. Other pipelines introduced triplane representations~\cite{zou2024triplane,xu2024agg} and canonical space generation~\cite{ye2024no} for feedforward 3D reconstruction and generation. SplatFormer~\cite{chen2024splatformer} instead uses a point transformer to refine an already optimized Gaussian scene for novel-view extrapolation. 

Pixel-wise prediction strategies, including multi-layered approaches like HiSplat~\cite{tang2024hisplat}, can generate redundant primitives in homogeneous regions. Recent methods explore alternatives: TokenGS~\cite{ren2026tokengs} uses learnable queries to decouple Gaussian count from image resolution and view count, while C3G~\cite{an2025c3g} predicts a compact Gaussian set from unposed multi-view images using learned queries. VolSplat~\cite{wang2026volsplat} lifts features using predicted depth and decodes Gaussians from occupied voxels. Off The Grid~\cite{moreau2026off} detects subpixel primitive positions and adapts patch density using image entropy. CompactSplat adopts a \textbf{content-aware allocation paradigm} based on an image-space quadtree. TAP concentrates finer partitions in textured regions, while HPS samples pooled Gaussian attributes and modulates primitive scales according to block footprints. This spatial hierarchy supports inference-time quality--compactness control with one checkpoint, reducing per-view redundancy while preserving detail at the selected operating point.

\section{Method}

\vspace{-5pt}
\subsection{Model Architecture}
\begin{figure}[h]
  \centering
  \includegraphics[width=1\linewidth]{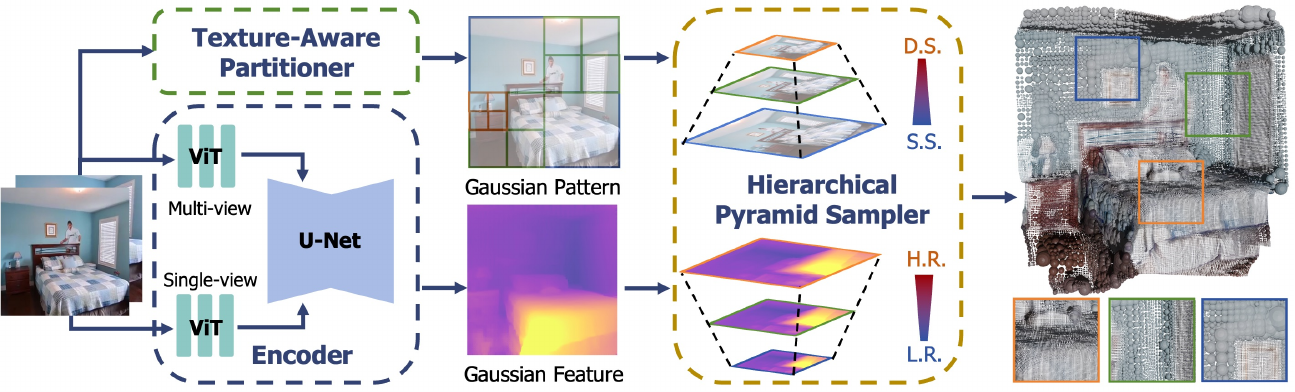}
    \caption{\textbf{Overview of CompactSplat.} A dual-path encoder first extracts and fuses multi-view and single-view features from sparse inputs, which are then processed by a multi-scale U-Net. Concurrently, the Texture-Aware Partitioner (TAP) evaluates scene complexity based on a user-defined threshold to generate multi-level spatial masks. Guided by these masks, the Hierarchical Pyramid Sampler (HPS) adaptively extracts features across decoder levels. Ultimately, this paradigm allocates sparse samples (S.S.) from low-resolution (L.R.) maps to flat regions, and dense samples (D.S.) from high-resolution (H.R.) maps to texturally complex areas.}
  \label{fig:model}
  \vspace{-15pt}
\end{figure}

Building upon the robust feature extraction paradigms established in recent feed-forward models~\cite{he2026surfsplat,xu2024depthsplat}, CompactSplat employs a dual-path encoder to capture both multi-view correspondences and single-view geometric priors. For the single-view branch, we utilize a pretrained monocular depth backbone (e.g., Depth Anything V2~\cite{yang2024depth}) to extract structural and semantic cues, which are particularly beneficial for ill-posed regions lacking cross-view overlaps. Concurrently, the multi-view branch processes the inputs through self- and cross-attention mechanisms~\cite{vaswani2017attention,liu2021swin} to construct plane-sweep cost volumes~\cite{chen2024mvsplat,collins1996space}, enforcing strict multi-view consistency. The complementary features from both branches are then concatenated to form a unified volumetric representation.

This fused feature volume is subsequently fed into a multi-scale 2D U-Net~\cite{ronneberger2015u} architecture. Unlike prior works that deterministically upsample all features into a single, full-resolution map for pixel-wise Gaussian prediction, our architecture decouples primitive generation to eliminate spatial redundancy, as illustrated in Fig.~\ref{fig:model}. Specifically, we introduce the Texture-Aware Partitioner (TAP), which analyzes spatial frequency and regional complexity from the inputs to produce a set of hierarchical partition masks. These masks dynamically dictate the appropriate Level of Detail (LoD) for representing different regions of the scene.

Guided by the TAP masks, our Hierarchical Pyramid Sampler (HPS) predicts explicit Gaussian attributes—including 3D center, scale, rotation, opacity, and spherical harmonics—directly from the multi-scale intermediate features of the U-Net decoder. Crucially, large, homogeneous regions are assigned to lower-resolution feature levels, allowing a single, scale-modulated Gaussian to seamlessly tile broad planar surfaces. In contrast, structurally complex areas are routed to the highest-resolution decoder features to preserve intricate geometric details. This multi-scale routing paradigm not only captures fine geometry but also naturally enables dynamic sparsity control, resulting in a highly compact yet expressive 3D representation.
\refstepcounter{mypara} \label{para:assumption}
The core motivation behind the Texture-Aware Partitioner (TAP) is to replace redundant pixel-wise generation with adaptive, density-aware primitive allocation. To this end, we formulate the image partitioning as a top-down quadtree decomposition driven by local textural complexity. This design is motivated by the inherent coupling of geometry and appearance within a single 3D Gaussian primitive. In natural environments, geometric complexity (e.g., depth discontinuities, occlusions, and structural boundaries) almost inevitably manifests as high-frequency textural variance. Although the converse is not strictly true—intricate textures frequently lie on structurally flat surfaces—relying on textural complexity as a subdivision proxy guarantees a conservative and sufficient upper bound. It ensures that all complex geometries receive dense primitive allocation, while aggressively pruning homogeneous regions where neither geometry nor texture demands high-resolution modeling.

\vspace{-5pt}
\subsection{Texture-Aware Partitioner (TAP)}
\vspace{-5pt}
\begin{wrapfigure}{r}{0.4\textwidth}
  \begin{minipage}{\linewidth}
    \vspace{-10pt}
    \rule{\linewidth}{0.8pt}
    \vspace{-15pt}

    \captionsetup[algorithm]{font=footnotesize, labelfont=bf}
    \captionof{algorithm}{Texture-Aware Partitioning}
    
    \vspace{-5pt}
    \rule{\linewidth}{0.5pt}
    \label{alg:tap}
    \scriptsize
    \setlength{\baselineskip}{1.4em}
    \algrenewcommand\alglinenumber[1]{\scriptsize\color{gray}#1:}
    \textbf{Input:} Input Image $I$, Partition Threshold $\tau$, Max Block Size $S_{max}$ \\
    \textbf{Output:} Partition Set $\mathcal{P} = \emptyset$
    
    \begin{algorithmic}[1]
    \vspace{2pt}
    \State Initialize queue $Q$ with initial blocks $B = (\boldsymbol{\mu}, \boldsymbol{b}, l)$, where $\boldsymbol{b} = S_{max} \times S_{max}$ and $l = 0$
    \State Initialize partition set $\mathcal{P} \gets \emptyset$
    \While{$Q$ is not empty}
        \State $B \gets Q\text{.pop()}$
        \State Compute complexity $\Delta(B)$
        \If{$\Delta(B) > \tau$ \textbf{and} $\boldsymbol{b} > 1 \times 1$}
            \State Split $B$ into 4 sub-quadrants $\{B_{child}\}$
            \State $Q\text{.push}(\{B_{child}\})$
        \Else
            \State $\mathcal{P} \gets \mathcal{P} \cup B$ 
        \EndIf
    \EndWhile
    \State \Return $\mathcal{P}$
    \end{algorithmic}
    \vspace{-5pt}
    \rule{\linewidth}{0.8pt} 
  \end{minipage}
  \vspace{-30pt}
\end{wrapfigure}

Given an input image $I \in \mathbb{R}^{C \times H \times W}$ and a predefined maximum block size $S_{\text{max}}$, TAP first initializes the image plane as a coarse grid of $S_{\text{max}} \times S_{\text{max}}$ patches. For each block $B$, we evaluate its internal textural homogeneity. Specifically, we define the intra-block complexity based on the maximum channel-wise intensity spread. Let $I_c(x,y)$ denote the pixel value at spatial coordinate $(x,y)$ for channel $c$. The splitting criterion for block $B$ is determined by calculating the maximum difference:
\begin{equation}
    \Delta(B) = \max_{c \in [1, C]} \left( \max_{(x,y) \in B} I_c(x,y) - \min_{(x,y) \in B} I_c(x,y) \right)
\end{equation}

To realize dynamic sparsity control, we introduce a user-defined partition threshold $\tau$. If the complexity $\Delta(B)$ exceeds $\tau$ and the block dimensions are strictly greater than $1 \times 1$, the region is considered texturally complex. Consequently, block $B$ is recursively subdivided into four equal-sized child quadrants. This recursive subdivision continues until the block satisfies $\Delta(B) \le \tau$ (indicating a smooth, homogeneous region) or reaches the single-pixel limit.

The resulting leaf nodes of this quadtree represent the final spatially adaptive partition. For each leaf node $k$, TAP outputs its normalized 2D center coordinate $\boldsymbol{\mu}_k \in [-1, 1]^2$, spatial dimensions $\boldsymbol{b}_k \in [1, S_{\text{max}}]^2$, and a Level of Detail (LoD) index $l_k \in [0, \log_2(S_{\text{max}}) - 1]$ corresponding to the depth of the subdivision. To improve robustness, especially near image boundaries and invalid view regions, we apply a binary mask to filter out non-informative nodes.

\textbf{This quadtree-based formulation naturally resolves spatial redundancy}: flat regions terminate early at lower LoD levels, yielding fewer and larger blocks, whereas complex high-frequency regions are further subdivided into finer blocks at higher LoD levels. Importantly, the threshold $\tau$ serves as a dynamic control knob: adjusting $\tau$ at inference time directly modulates the overall primitive count without requiring retraining. The resulting tuples $\{(\boldsymbol{\mu}_k, \boldsymbol{b}_k, l_k)\}$ are then passed to the Hierarchical Pyramid Sampler to guide feature extraction.
\textbf{This quadtree-based formulation naturally resolves spatial redundancy}: flat regions terminate early at lower LoD levels, yielding fewer and larger blocks, whereas complex high-frequency regions are further subdivided into finer blocks at higher LoD levels.




\subsection{Hierarchical Pyramid Sampler (HPS)}

\begin{wrapfigure}{r}{0.4\textwidth}
  \begin{minipage}{\linewidth}
    \vspace{-10pt}
    \rule{\linewidth}{0.8pt}
    \vspace{-15pt}
    \captionsetup[algorithm]{font=footnotesize, labelfont=bf}
    \captionof{algorithm}{Hierarchical Pyramid Sampler}
    \vspace{-5pt}
    \rule{\linewidth}{0.5pt}
    \label{alg:tap}
    \scriptsize
    \setlength{\baselineskip}{1.4em}
    \algrenewcommand\alglinenumber[1]{\scriptsize\color{gray}#1:}
\textbf{Input:} Base attribute map $\mathbf{A}_{base}$, Partition Set $\mathcal{P}$, Max pyramid levels $L$ \\
\textbf{Output:} 3D Gaussians Set $\mathcal{G} = \emptyset$
\begin{algorithmic}[1]
\vspace{2pt}
\State \textit{\% 1. Construct Attribute Scale Space (Pyramid)}
\State $\mathcal{A}[L-1] \gets \mathbf{A}_{base}$
\For{$l = L-1$ \textbf{to} $0$}
    \State $\mathcal{A}[l-1] \gets \text{AveragePool}(\mathcal{A}[l])$ 
\EndFor

\vspace{2pt}
\State \textit{\% 2. Hierarchical Pyramid Sampling}
\For{\textbf{each} $(\boldsymbol{\mu}_k, \mathbf{b}_k, l_k) \in \mathcal{P}$}
    \State $\mathbf{a}_k \gets \text{GridSample}(\mathcal{A}[l_k], \boldsymbol{\mu}_k)$ 
    \State $\boldsymbol{\mu}_k, \mathbf{r}_k, \Delta\mathbf{s}_k, \boldsymbol{\alpha}_k, \mathbf{c}_k \gets \text{Decode}(\mathbf{a}_k)$
    \State $\mathbf{s}_k \gets \Delta\mathbf{s}_k \odot \mathbf{b}_k$ 
    \State $\mathcal{G} \gets \mathcal{G} \cup \{ \text{Gaussian}(\boldsymbol{\mu}_k, \mathbf{s}_k, \mathbf{r}_k, \boldsymbol{\alpha}_k, \mathbf{c}_k) \}$
\EndFor

\State \Return $\mathcal{G}$
    \end{algorithmic}
    \vspace{-5pt}
    \rule{\linewidth}{0.8pt} 
  \end{minipage}
  \vspace{-15pt}
\end{wrapfigure}

While the Texture-Aware Partitioner (TAP) determines the optimal spatial layout and density of the primitives, the Hierarchical Pyramid Sampler (HPS) maps these abstract spatial partitions into explicit 3D Gaussian attributes. To this end, HPS performs scale-conditioned sampling from a foundational scene representation.

Instead of forcing the network to implicitly predict multi-scale attributes across different decoder layers, our architecture utilizes a DPT head to predict a unified, high-fidelity Gaussian attribute field $\mathbf{A}_{\text{base}}$ at full image resolution. However, directly sampling from this high-frequency field for large, homogeneous partitions would inherently lead to aliasing artifacts and a mismatch in the receptive field. To address this, we construct a discrete attribute scale space $\mathcal{A} = \{\mathbf{A}_{L-1}, \dots, \mathbf{A}_1, \mathbf{A}_0\}$. Starting with the full-resolution field $\mathbf{A}_{L-1} = \mathbf{A}_{\text{base}}$, each subsequent coarser level $\mathbf{A}_l$ is derived by progressively applying a spatial low-pass filter (e.g., local aggregation). This multi-scale representation ensures that features at lower Level of Detail (LoD) indices naturally encapsulate broader regional contexts, perfectly mirroring the TAP quadtree hierarchy.

Given a quadtree leaf node $k$ from TAP, parameterized by its normalized center $\boldsymbol{\mu}_k \in [-1, 1]^2$, spatial footprint $\boldsymbol{b}_k$, and LoD index $l_k$, HPS performs discrete-to-continuous feature querying. By routing the sampling operation to the corresponding scale space level $\mathbf{A}_{l_k}$, the extracted attribute vector $\mathbf{a}_k$ is intrinsically aligned with the spatial frequency of the partition:

\vspace{-10pt}
\begin{equation}
\mathbf{a}_k = \text{GridSample}(\mathbf{A}_{l_k}, \boldsymbol{\mu}_k)
\end{equation}
\vspace{-10pt}

A central design choice of HPS is its scale-aware geometric decoding. The queried vector $\mathbf{a}_k$ encapsulates base properties such as opacity, rotation, spherical harmonics, and a relative scale multiplier $\Delta\mathbf{s}_k$. To guarantee that the allocated primitive physically spans its designated spatial partition, the final scale of the Gaussian, denoted as $\mathbf{s}_k$, is explicitly modulated by the quadtree node dimensions:

\vspace{-10pt}
\begin{equation}
\mathbf{s}_k = \Delta\mathbf{s}_k \odot \boldsymbol{b}_k
\end{equation}
\vspace{-10pt}

This explicit scale conditioning elegantly bridges the gap between 2D partition logic and 3D geometry. It enables a single Gaussian, sampled from a deeper level of the scale space, to morphologically expand and seamlessly tile expansive planar surfaces. This fundamentally breaks the representational redundancy of rigid pixel-wise predictions while rigorously preserving the rendering quality of high-frequency intricate details. Moren details are provided in the appendix.

\vspace{-5pt}
\subsection{Training Strategy}
\vspace{-5pt}
A core objective of CompactSplat is to establish a unified paradigm for controllable inference, enabling dynamic trade-offs between rendering fidelity and sparsity without network retraining. To achieve this, the model must be exposed to varying partition thresholds $\tau$ during the training process. However, jointly optimizing for multi-view geometric consistency and dynamic primitive pruning from scratch introduces severe training instability. To mitigate this, we propose a tailored two-stage curriculum learning strategy.

\textbf{Stage 1: Dense Geometric Initialization.} In the initial phase, we strictly fix the partition threshold to $\tau = 0$. Under this condition, the Texture-Aware Partitioner (TAP) behaves conservatively, decomposing the feature maps at the highest Level of Detail (LoD) across the entire scene. This dense supervision isolates the learning objective, forcing the dual-path encoder and the multi-scale U-Net to focus exclusively on establishing robust multi-view correspondences and extracting accurate geometric priors, completely free from the confounding effects of adaptive sparsity.

\textbf{Stage 2: Sparsity-Aware Fine-Tuning.} Once the network has converged to a reliable geometric understanding, we introduce stochastic thresholding. In this subsequent stage, $\tau$ is randomly sampled from a predefined uniform distribution $\tau \sim \mathcal{U}(0, \tau_{\text{max}})$ at each training iteration. This strategy exposes the model to a continuous spectrum of pruning ratios. Crucially, it forces the Hierarchical Pyramid Sampler (HPS) to accurately learn the geometric scale multipliers for lower-resolution Gaussians, ensuring they can morphologically expand to seamlessly tile broad, homogeneous regions under aggressive pruning scenarios. Through this two-stage curriculum, a single set of pretrained weights inherently acquires the versatility to reconstruct scenes at arbitrary sparsity levels on the fly.

\vspace{-5pt}
\subsection{Training Loss}
\vspace{-5pt}
To optimize our feed-forward architecture, we apply direct photometric supervision between the synthesized views and the corresponding target images. The overall objective function is formulated as a joint optimization of pixel-level fidelity and perceptual quality. Specifically, for a training batch containing $M$ samples, the loss is defined as:

\begin{equation}
L_{\mathrm{total}} = \sum_{m=1}^{M} \left( 
\operatorname{MSE}\left(I_{\mathrm{render}}^m, I_{\mathrm{gt}}^m\right) 
+ \lambda \cdot \operatorname{LPIPS}\left(I_{\mathrm{render}}^m, I_{\mathrm{gt}}^m\right) 
\right),
\end{equation}

where $I_{\mathrm{render}}^m$ and $I_{\mathrm{gt}}^m$ denote the $m$-th rendered output and its corresponding ground truth, respectively. Consistent with established training protocols in recent generalizable 3DGS frameworks~\cite{charatan2024pixelsplat,chen2024mvsplat,xu2024depthsplat,he2026surfsplat}, the scaling factor $\lambda$ for the perceptual loss is empirically set to 0.05.

\vspace{-10pt}
\section{Experiment}
\vspace{-8pt}

\subsection{Main Results}

\begin{figure}[ht]
  \vspace{-12pt}
  \centering
  \includegraphics[width=1\linewidth]{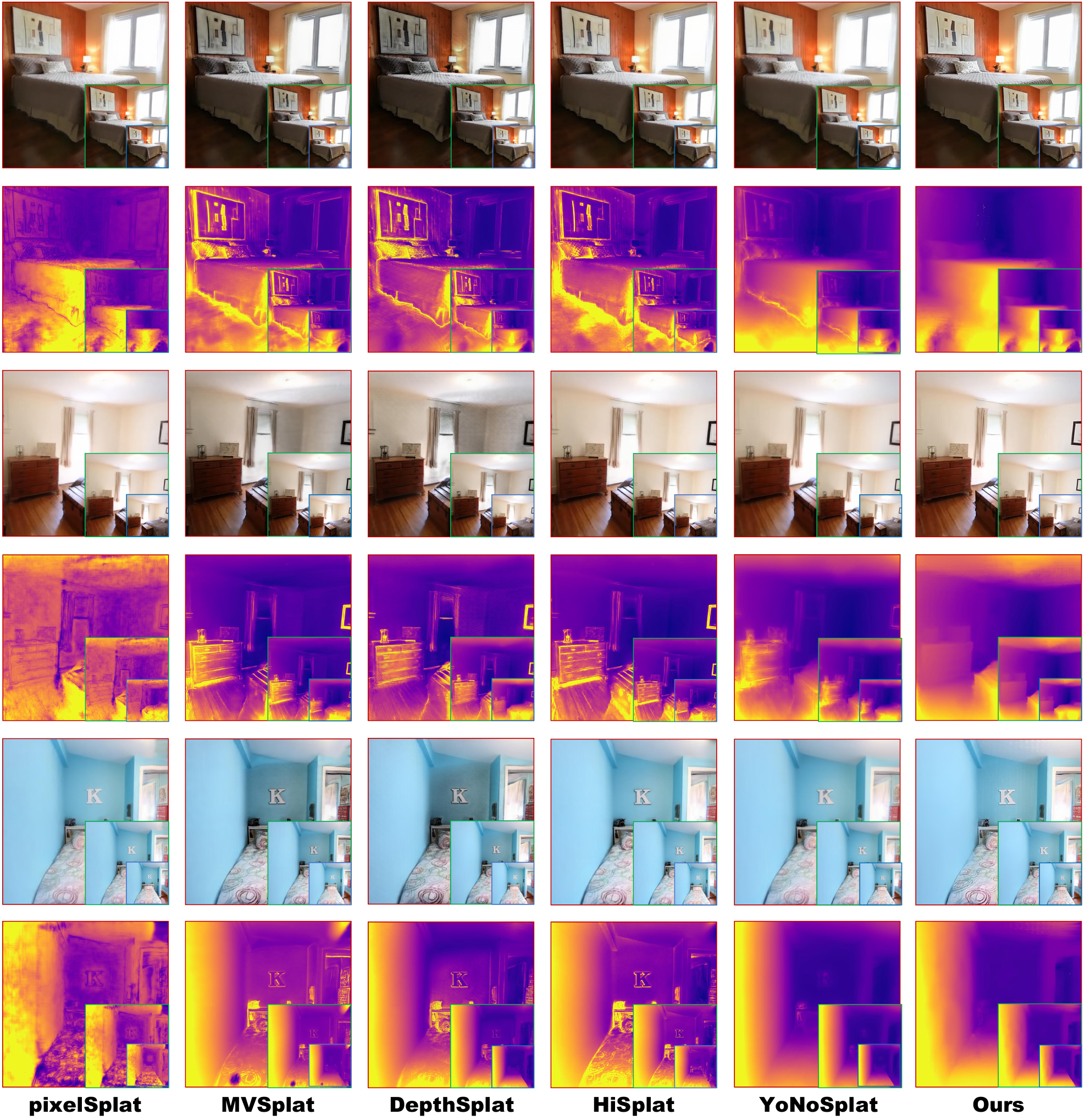}
  \caption{\textbf{Multi-resolution rendering comparison of 3D scenes.} 
    We visualize rendered RGB images and depth maps at three scaling factors: $\times1$ ({\color{blue!70}blue} box), $\times2$ ({\color{green!60!black}green} box), and $\times4$ ({\color{red!70}red} box). 
    Under higher magnification, structural deficiencies in conventional 3D representations become increasingly apparent. 
    In the RGB domain, primitive sparsity causes unfilled gaps, manifesting as dark, hollow regions. 
    Correspondingly, in the depth space (where yellow indicates near and blue indicates distant surfaces), these geometric discontinuities erroneously register as spurious near-field artifacts. 
    In contrast, CompactSplat's scale-modulated primitives seamlessly tile the surfaces, yielding watertight geometry and artifact-free renderings across all resolutions.}
  \label{fig:mainresult}
  \vspace{-10pt}
\end{figure}

\refstepcounter{mypara} \label{para:expsta}
\begin{table}[ht]
\centering
\vspace{-5pt}
\caption{Novel view synthesis performance on the RealEstate10k dataset.}
\label{re10k}
\fontsize{6}{9.5}\selectfont
\setlength{\tabcolsep}{3.1pt}
\begin{tabular}{lccc|ccc|ccc|ccc|cc}
\toprule
& \multicolumn{3}{c|}{256$\times$256 (\textbf{Standard})} & \multicolumn{3}{c|}{512$\times$512 (\textbf{HRRC})} & \multicolumn{3}{c|}{1024$\times$1024 (\textbf{HRRC})} & \multicolumn{3}{c|}{Average} & \multicolumn{2}{c}{CPT} \\
Method & PSNR↑ & SSIM↑ & LPIPS↓ & PSNR↑ & SSIM↑ & LPIPS↓ & PSNR↑ & SSIM↑ & LPIPS↓ & PSNR↑ & SSIM↑ & LPIPS↓ & GPP↓ & GS Num↓\\
\midrule
pixelSplat~\citeyear{charatan2024pixelsplat}  & 26.049 & 0.862 & 0.137 & 25.782 & 0.868 & 0.207 & 24.920 & 0.877 & 0.269 & 25.584 & 0.869 & 0.204 & 3.000 & 393.2k  \\
HiSplat~\citeyear{tang2024hisplat}  & 27.193 & 0.882 & 0.117 & 25.269 & 0.870 & 0.198 & 24.262 & 0.878 & 0.248 & 25.575 & 0.877 & 0.188 & 1.750 & 229.4k  \\

\midrule
MVSplat~\citeyear{chen2024mvsplat} & 26.359 & 0.868 & 0.129 & 20.408 & 0.809 & 0.290 & 17.966 & 0.755 & 0.425 & 21.578 & 0.811 & 0.281 & 1.000 & 131.1k  \\
TranSplat~\citeyear{zhang2025transplat} & 26.687 & 0.875 & 0.125 & 20.610 & 0.815 & 0.286 & 18.154 & 0.761 & 0.427 & 21.817 & 0.817 & 0.279 & 1.000 & 131.1k \\
DepthSplat~\citeyear{xu2024depthsplat} & 27.504 & 0.890 & \textbf{0.112} & 20.031 & 0.774 & 0.341 & 16.385 & 0.635 & 0.491 & 21.307 & 0.766 & 0.315 & 1.000 & 131.1k \\
SurfSplat~\citeyear{he2026surfsplat} & \underline{27.537} & 0.892 & \textbf{0.112} & 26.331 & 0.866 & 0.217 & 24.897 & 0.842 & 0.320 & 26.255 & 0.867 & 0.216 & 1.000 & 131.1k \\
YonoSplat$^*$~\citeyear{ye2025yonosplatneedmodelfeedforward} & 25.735 & 0.851 & 0.130 & 25.415 & 0.857 & \underline{0.205} & 24.785 & \textbf{0.872} & \textbf{0.261} & 25.312 & 0.860 & \underline{0.199} & 1.000 & 131.1k \\
Ours-S & 27.086 & 0.884 & 0.123 & 26.066 & 0.866 & \underline{0.205} & 25.118 & 0.857 & \underline{0.276} & 26.090 & 0.869 & 0.201 & \textbf{0.554} & \textbf{72.6k}  \\
Ours-B & 27.523 & \underline{0.891} & 0.118 & \underline{26.332} & \underline{0.870} & \textbf{0.202} & \underline{25.240} & \underline{0.858} & \underline{0.276} & \underline{26.365} & \underline{0.873} & \underline{0.199} & \textbf{0.554} & \textbf{72.6k}  \\
Ours-L & \textbf{27.567} & \textbf{0.892} & \underline{0.117} & \textbf{26.357} & \textbf{0.871} & \textbf{0.202} & \textbf{25.252} & \underline{0.858} & \underline{0.276} & \textbf{26.392} & \textbf{0.874} & \textbf{0.198} & \textbf{0.554} & \textbf{72.6k}  \\
\bottomrule
\end{tabular}

\begin{flushleft}
    \footnotesize $^*$Note: Due to architectural constraints, YonoSplat is evaluated at $224\times224$ resolution, which may yield marginally inflated metrics compared to the standard $256\times256$ resolution used by other baselines.
\end{flushleft}
\vspace{-25pt}
\end{table}

\begin{table}[ht]
\centering
\caption{Novel view synthesis performance on the ACID dataset.}
\label{acid}
\fontsize{6}{9.5}\selectfont
\setlength{\tabcolsep}{3.3pt}
\begin{tabular}{lccc|ccc|ccc|ccc|cc}
\toprule
& \multicolumn{3}{c|}{256$\times$256 (\textbf{Standard})} & \multicolumn{3}{c|}{512$\times$512 (\textbf{HRRC})} & \multicolumn{3}{c|}{1024$\times$1024 (\textbf{HRRC})} & \multicolumn{3}{c}{Average} & \multicolumn{2}{c}{CPT} \\
Method & PSNR↑ & SSIM↑ & LPIPS↓ & PSNR↑ & SSIM↑ & LPIPS↓ & PSNR↑ & SSIM↑ & LPIPS↓ & PSNR↑ & SSIM↑ & LPIPS↓ & GPP↓ & GS Num↓\\
\midrule
pixelSplat~\citeyear{charatan2024pixelsplat}  & 28.284 & 0.842 & 0.146 & 27.687 & 0.848 & 0.243 & 26.462 & 0.858 & 0.343 & 27.478 & 0.849 & 0.244 & 3.000 & 393.2k \\
HiSplat~\citeyear{tang2024hisplat}  & 28.737 & 0.853 & 0.132 & 25.376 & 0.833 & 0.246 & 23.988 & 0.841 & 0.314 & 25.700 & 0.842 & 0.231 & 1.750 & 229.4k\\
\midrule
MVSplat~\citeyear{chen2024mvsplat} & 28.202 &\underline;{0.842} & 0.145 & 17.802 & 0.711 & 0.406 & 14.784 & 0.572 & 0.567 & 20.263 & 0.708 & 0.373 & 1.000 & 131.1k \\
TranSplat~\citeyear{zhang2025transplat} & \textbf{28.337} & \textbf{0.845} & \textbf{0.143} & 17.911 & 0.716 & 0.402 & 14.956 & 0.582 & 0.558 & 20.401 & 0.714 & 0.373 & 1.000 & 131.1k \\
SurfSplat~\citeyear{he2026surfsplat} & \underline{28.336} & \textbf{0.845} & \underline{0.144} & \textbf{26.868} & \textbf{0.814} & \underline{0.281} & \underline{21.253} & \underline{0.690} & \underline{0.457} & \underline{25.486} & \underline{0.783} & \underline{0.294} & 1.000 & 131.1k \\
Ours & 27.992 & 0.837 & 0.157 & \underline{26.384} & \underline{0.810} & \textbf{0.271} & \textbf{24.063} & \textbf{0.782} & \textbf{0.376} & \textbf{26.146} & \textbf{0.810} & \textbf{0.268} & \textbf{0.625} & \textbf{91.4k} \\
\bottomrule
\end{tabular}
\vspace{-25pt}
\end{table}

\noindent\textbf{Datasets.}To comprehensively evaluate our approach, we benchmark CompactSplat across a diverse set of environments commonly utilized in generalizable 3DGS literature. Specifically, we conduct experiments on RealEstate10K (RE10K)~\cite{zhou2018stereo} for indoor scenes and ACID~\cite{liu2021infinite} for outdoor aerial trajectories, strictly adhering to the official train-test splits and precomputed camera poses established by prior works. To further validate the cross-dataset generalization and the robustness of our context-aware partitioning on complex geometries, we extend our evaluation to the DTU~\cite{jensen2014large}, DL3DV~\cite{ling2024dl3dv}, and the highly challenging ScanNet~\cite{dai2017scannet} datasets. Comprehensive cross-dataset evaluations are detailed in the supplementary material.

\noindent\textbf{Evaluation Metrics.} We quantify the fidelity of novel view synthesis using standard image quality assessment metrics: Peak Signal-to-Noise Ratio (PSNR), Structural Similarity Index (SSIM), and Learned Perceptual Image Patch Similarity (LPIPS). Furthermore, to validate the structural integrity and watertight properties of our representations, we employ the High-Resolution Rendering Consistency (HRRC) metric~\cite{he2026surfsplat} at both $2\times$ and $4\times$ spatial magnifications. Crucially, to demonstrate the efficiency of our feedforward architecture, we introduce two representation-specific metrics: the average \textbf{Number of Gaussians(GS Num.)}, and the \textbf{Gaussian Per Pixel(GPP)}, which explicitly highlights the \textbf{compactness(CPT)} of primitives compared to the unpruned baseline.

\noindent\textbf{Baselines.}Our framework is compared against leading sparse-view feedforward 3DGS models, including PixelSplat~\cite{charatan2024pixelsplat}, MVSplat~\cite{chen2024mvsplat}, TranSplat~\cite{zhang2025transplat}, HiSplat~\cite{tang2024hisplat}, DepthSplat~\cite{xu2024depthsplat}, SurfSplat~\cite{he2026surfsplat} and YonoSplat~\cite{ye2025yonosplatneedmodelfeedforward}. While architectures like PixelSplat and HiSplat indiscriminately predict multiple Gaussians per pixel, methods such as MVSplat, TranSplat, DepthSplat, SurfSplat and YonoSplat restrict their generation to a single primitive per pixel. Because our core contribution lies in decisively breaking the 1:1 pixel-to-primitive barrier to eliminate spatial redundancy, benchmarking against this latter group establishes a highly rigorous baseline for evaluating representational efficiency.

\noindent\textbf{Qualitative Comparisons.} 
Visual inspections of the rendered novel views in Figure~\ref{fig:mainresult} demonstrate that CompactSplat achieves photorealistic quality that is highly competitive with, and frequently surpasses, current state-of-the-art feed-forward methods. This superiority becomes particularly evident under high-resolution magnification. While baseline models often struggle with spatial voids, noisy boundaries, and spurious artifacts due to their rigid primitive allocation, CompactSplat maintains strict structural integrity. By leveraging complexity-aware allocation, our method seamlessly tiles flat planar surfaces with scale-modulated primitives while accurately preserving high-frequency edges. Consequently, it yields strictly watertight geometries and visually coherent structures that remain exceptionally robust even upon close inspection. Additional qualitative comparisons across various scenes are provided in our \textbf{supplementary video}.

\noindent\textbf{Quantitative Results.} 
The objective metrics presented in Table~\ref{re10k} and Table~\ref{acid} further corroborate the transformative efficiency of our architecture. Across standard fidelity metrics (e.g., PSNR, SSIM, LPIPS) and the structure-aware HRRC metric, CompactSplat consistently matches or exceeds the performance of leading baselines. Crucially, it achieves this high-fidelity synthesis while drastically reducing the total Number of Gaussians. By pioneering a departure from the conventional pixel-wise prediction paradigm, CompactSplat establishes a highly compact representation. Furthermore, this compactness is inherently controllable; our architecture allows users to dynamically modulate the pruning ratio on the fly, offering an unprecedented capability to balance rendering quality and memory footprint without requiring any network retraining.

\vspace{-10pt}
\subsection{Analysis of Dynamic Sparsity Control}
\vspace{-5pt}


A core capability of CompactSplat is its inherent support for dynamic sparsity control without requiring network retraining. To investigate the efficacy of this mechanism, we systematically analyze the trade-off between reconstruction quality and representational compactness by modulating the partition threshold $\tau$ during inference, which effectively controls the average Gaussians Per Pixel.

\begin{wrapfigure}{l}{0.55\linewidth}
\vspace{-15pt}
\includegraphics[width=\linewidth]{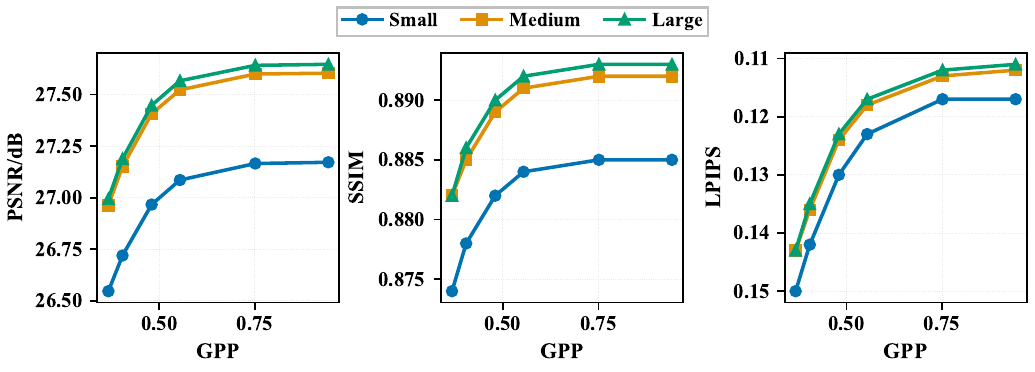}
\captionsetup{belowskip=0pt}
\caption{\textbf{Quantitative analysis.} 
By modulating the partition threshold $\tau$ at inference time, CompactSplat continuously traverses the Pareto frontier. The model achieves extreme reductions in primitive count (measured in Gaussian per pixel, GPP) with virtually negligible degradation in rendering fidelity.}
\label{fig:ablation_quantitative}


\includegraphics[width=\linewidth]{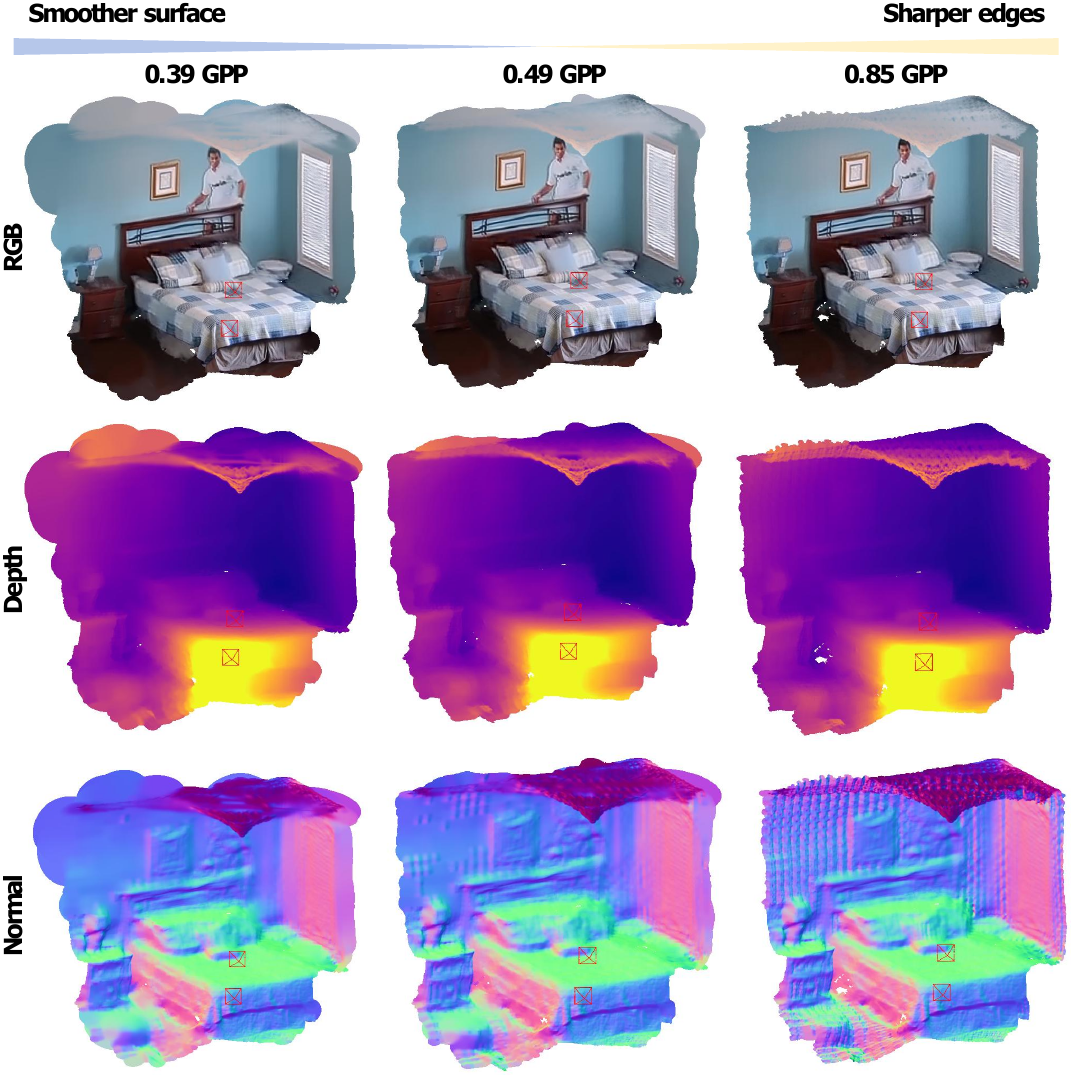}
\captionsetup{belowskip=-18pt}
\caption{\textbf{Qualitative analysis.} 
Visualizations under varying GPP settings reveal an elegant architectural trade-off. Lower GPP limits enforce the use of expanded, scale-modulated primitives, yielding smoother, watertight planar surfaces. Conversely, higher GPP settings allocate denser primitives to more faithfully recover high-frequency edges and complex boundaries.}
\label{fig:ablation_qualitative}
\end{wrapfigure}

\noindent\textbf{Quantitative Trade-offs.} We evaluate our three model variants (Ours-S, Ours-B, and Ours-L) across a continuous spectrum of GPP constraints (see Fig.~\ref{fig:ablation_quantitative}). As demonstrated by the objective metrics, CompactSplat exhibits a highly resilient Pareto frontier. Modulating the threshold to reduce the GPP within a reasonable operational range incurs only a marginal degradation in standard fidelity metrics. In exchange for this negligible drop in quantitative performance, our method yields substantial gains in representational compactness. This asymmetric trade-off confirms that the model successfully eliminates spatial redundancy, offering users a highly flexible knob to drastically reduce memory footprint and rendering overhead while rigorously preserving overall scene integrity.

\noindent\textbf{Qualitative Observations.} This robust quantitative performance is further corroborated by visual inspections of the rendered RGB, depth, and normal maps across varying GPP configurations (see Fig.~\ref{fig:ablation_qualitative}). We observe distinct and highly complementary geometric behaviors at opposite ends of the sparsity spectrum. Operating at a high GPP allows the network to allocate dense, high-resolution primitives along intricate structures, yielding exceptionally sharp edges and crisp high-frequency boundaries. Conversely, operating at a lower GPP forces the architecture to leverage larger, scale-modulated primitives sampled from deeper pyramid levels. Interestingly, this highly pruned setting is intrinsically suited for modeling broad, homogeneous planar surfaces (e.g., walls and floors). By seamlessly tiling these expansive areas with single, expanded Gaussians, the low-GPP configuration fundamentally eliminates the spatial voids and "hollow" artifacts common in traditional sparse-view 3DGS, resulting in strictly watertight and visually coherent structural reconstructions.
\vspace{-12pt}
\section{Conclusion}
\vspace{-7pt}

In this paper, we presented \textbf{CompactSplat}, a novel feed-forward framework designed to overcome the severe spatial redundancy inherent in current sparse-view 3D Gaussian Splatting models. By synergizing a Texture-Aware Partitioner (TAP) with a Hierarchical Pyramid Sampler (HPS), our approach decisively breaks away from the rigid pixel-wise prediction paradigm, adaptively allocating primitives based on regional complexity. This scale-aware representation not only achieves state-of-the-art rendering fidelity across multiple datasets but also inherently supports dynamic sparsity control during inference. Consequently, CompactSplat delivers highly compact 3D priors, offering flexible, on-the-fly trade-offs between reconstruction quality and memory footprint.

\newpage
\section*{Acknowledgement}
\vspace{-5pt}
This work was partly supported by the NSFC (62431015, 62571317, 62501387), the Fundamental Research Funds for the Central Universities, Shanghai Key Laboratory of Digital Media Processing and Transmission under Grant 22DZ2229005, 111 project BP0719010.

{
    \small
    \bibliographystyle{neurips_2025}
    \bibliography{neurips_2025}

@article{mildenhall2021nerf,
  title={Nerf: Representing scenes as neural radiance fields for view synthesis},
  author={Mildenhall, Ben and Srinivasan, Pratul P and Tancik, Matthew and Barron, Jonathan T and Ramamoorthi, Ravi and Ng, Ren},
  journal={Communications of the ACM},
  volume={65},
  number={1},
  pages={99--106},
  year={2021},
  publisher={ACM New York, NY, USA}
}

@inproceedings{fridovich2022plenoxels,
  title={Plenoxels: Radiance fields without neural networks},
  author={Fridovich-Keil, Sara and Yu, Alex and Tancik, Matthew and Chen, Qinhong and Recht, Benjamin and Kanazawa, Angjoo},
  booktitle={Proceedings of the IEEE/CVF Conference on Computer Vision and Pattern Recognition},
  pages={5501--5510},
  year={2022}
}

@article{muller2022instant,
  title={Instant neural graphics primitives with a multiresolution hash encoding},
  author={M{\"u}ller, Thomas and Evans, Alex and Schied, Christoph and Keller, Alexander},
  journal={ACM transactions on graphics (TOG)},
  volume={41},
  number={4},
  pages={1--15},
  year={2022},
  publisher={ACM New York, NY, USA}
}

@inproceedings{barron2021mip,
  title={Mip-nerf: A multiscale representation for anti-aliasing neural radiance fields},
  author={Barron, Jonathan T and Mildenhall, Ben and Tancik, Matthew and Hedman, Peter and Martin-Brualla, Ricardo and Srinivasan, Pratul P},
  booktitle={Proceedings of the IEEE/CVF International Conference on Computer Vision},
  pages={5855--5864},
  year={2021}
}

@inproceedings{barron2022mip,
  title={Mip-nerf 360: Unbounded anti-aliased neural radiance fields},
  author={Barron, Jonathan T and Mildenhall, Ben and Verbin, Dor and Srinivasan, Pratul P and Hedman, Peter},
  booktitle={Proceedings of the IEEE/CVF Conference on Computer Vision and Pattern Recognition},
  pages={5470--5479},
  year={2022}
}

@article{kerbl20233d,
  title={3d gaussian splatting for real-time radiance field rendering},
  author={Kerbl, Bernhard and Kopanas, Georgios and Leimk{\"u}hler, Thomas and Drettakis, George},
  journal={ACM Transactions on Graphics},
  volume={42},
  number={4},
  pages={1--14},
  year={2023},
  publisher={ACM}
}

@article{luiten2023dynamic,
  title={Dynamic 3d gaussians: Tracking by persistent dynamic view synthesis},
  author={Luiten, Jonathon and Kopanas, Georgios and Leibe, Bastian and Ramanan, Deva},
  journal={arXiv preprint arXiv:2308.09713},
  year={2023}
}

@article{wu20234d,
  title={4d gaussian splatting for real-time dynamic scene rendering},
  author={Wu, Guanjun and Yi, Taoran and Fang, Jiemin and Xie, Lingxi and Zhang, Xiaopeng and Wei, Wei and Liu, Wenyu and Tian, Qi and Wang, Xinggang},
  journal={arXiv preprint arXiv:2310.08528},
  year={2023}
}

@article{huang2023sc,
  title={SC-GS: Sparse-Controlled Gaussian Splatting for Editable Dynamic Scenes},
  author={Huang, Yi-Hua and Sun, Yang-Tian and Yang, Ziyi and Lyu, Xiaoyang and Cao, Yan-Pei and Qi, Xiaojuan},
  journal={arXiv preprint arXiv:2312.14937},
  year={2023}
}

@article{he2024s4d,
  title={S4d: Streaming 4d real-world reconstruction with gaussians and 3d control points},
  author={He, Bing and Chen, Yunuo and Lu, Guo and Wang, Qi and Gu, Qunshan and Xie, Rong and Song, Li and Zhang, Wenjun},
  journal={arXiv preprint arXiv:2408.13036},
  year={2024}
}

@article{wan2024superpoint,
  title={Superpoint Gaussian Splatting for Real-Time High-Fidelity Dynamic Scene Reconstruction},
  author={Wan, Diwen and Lu, Ruijie and Zeng, Gang},
  journal={arXiv preprint arXiv:2406.03697},
  year={2024}
}

@article{paszke2019pytorch,
  title={Pytorch: An imperative style, high-performance deep learning library},
  author={Paszke, Adam and Gross, Sam and Massa, Francisco and Lerer, Adam and Bradbury, James and Chanan, Gregory and Killeen, Trevor and Lin, Zeming and Gimelshein, Natalia and Antiga, Luca and others},
  journal={Advances in neural information processing systems},
  volume={32},
  year={2019}
}

@inproceedings{schoenberger2016sfm,
    author={Sch\"{o}nberger, Johannes Lutz and Frahm, Jan-Michael},
    title={Structure-from-Motion Revisited},
    booktitle={Conference on Computer Vision and Pattern Recognition (CVPR)},
    year={2016},
}

@inproceedings{cheng2024gaussianpro,
  title={Gaussianpro: 3d gaussian splatting with progressive propagation},
  author={Cheng, Kai and Long, Xiaoxiao and Yang, Kaizhi and Yao, Yao and Yin, Wei and Ma, Yuexin and Wang, Wenping and Chen, Xuejin},
  booktitle={Forty-first International Conference on Machine Learning},
  year={2024}
}

@inproceedings{zhang2024fregs,
  title={Fregs: 3d gaussian splatting with progressive frequency regularization},
  author={Zhang, Jiahui and Zhan, Fangneng and Xu, Muyu and Lu, Shijian and Xing, Eric},
  booktitle={Proceedings of the IEEE/CVF Conference on Computer Vision and Pattern Recognition},
  pages={21424--21433},
  year={2024}
}

@article{radl2024stopthepop,
  title={Stopthepop: Sorted gaussian splatting for view-consistent real-time rendering},
  author={Radl, Lukas and Steiner, Michael and Parger, Mathias and Weinrauch, Alexander and Kerbl, Bernhard and Steinberger, Markus},
  journal={ACM Transactions on Graphics (TOG)},
  volume={43},
  number={4},
  pages={1--17},
  year={2024},
  publisher={ACM New York, NY, USA}
}

@inproceedings{diolatzis2024n,
  title={N-dimensional gaussians for fitting of high dimensional functions},
  author={Diolatzis, Stavros and Zirr, Tobias and Kuznetsov, Alexander and Kopanas, Georgios and Kaplanyan, Anton},
  booktitle={ACM SIGGRAPH 2024 Conference Papers},
  pages={1--11},
  year={2024}
}

@inproceedings{yan2024multi,
  title={Multi-scale 3d gaussian splatting for anti-aliased rendering},
  author={Yan, Zhiwen and Low, Weng Fei and Chen, Yu and Lee, Gim Hee},
  booktitle={Proceedings of the IEEE/CVF Conference on Computer Vision and Pattern Recognition},
  pages={20923--20931},
  year={2024}
}

@inproceedings{yu2024mip,
  title={Mip-splatting: Alias-free 3d gaussian splatting},
  author={Yu, Zehao and Chen, Anpei and Huang, Binbin and Sattler, Torsten and Geiger, Andreas},
  booktitle={Proceedings of the IEEE/CVF conference on computer vision and pattern recognition},
  pages={19447--19456},
  year={2024}
}

@article{song2024sa,
  title={Sa-gs: Scale-adaptive gaussian splatting for training-free anti-aliasing},
  author={Song, Xiaowei and Zheng, Jv and Yuan, Shiran and Gao, Huan-ang and Zhao, Jingwei and He, Xiang and Gu, Weihao and Zhao, Hao},
  journal={arXiv preprint arXiv:2403.19615},
  year={2024}
}

@inproceedings{liang2024analytic,
  title={Analytic-splatting: Anti-aliased 3d gaussian splatting via analytic integration},
  author={Liang, Zhihao and Zhang, Qi and Hu, Wenbo and Zhu, Lei and Feng, Ying and Jia, Kui},
  booktitle={European conference on computer vision},
  pages={281--297},
  year={2024},
  organization={Springer}
}

@inproceedings{girish2024eagles,
  title={Eagles: Efficient accelerated 3d gaussians with lightweight encodings},
  author={Girish, Sharath and Gupta, Kamal and Shrivastava, Abhinav},
  booktitle={European Conference on Computer Vision},
  pages={54--71},
  year={2024},
  organization={Springer}
}

@inproceedings{navaneet2024compgs,
  title={Compgs: Smaller and faster gaussian splatting with vector quantization},
  author={Navaneet, KL and Pourahmadi Meibodi, Kossar and Abbasi Koohpayegani, Soroush and Pirsiavash, Hamed},
  booktitle={European Conference on Computer Vision},
  pages={330--349},
  year={2024},
  organization={Springer}
}

@inproceedings{niedermayr2024compressed,
  title={Compressed 3d gaussian splatting for accelerated novel view synthesis},
  author={Niedermayr, Simon and Stumpfegger, Josef and Westermann, R{\"u}diger},
  booktitle={Proceedings of the IEEE/CVF Conference on Computer Vision and Pattern Recognition},
  pages={10349--10358},
  year={2024}
}

@inproceedings{lee2024compact,
  title={Compact 3d gaussian representation for radiance field},
  author={Lee, Joo Chan and Rho, Daniel and Sun, Xiangyu and Ko, Jong Hwan and Park, Eunbyung},
  booktitle={Proceedings of the IEEE/CVF Conference on Computer Vision and Pattern Recognition},
  pages={21719--21728},
  year={2024}
}

@article{fan2024lightgaussian,
  title={Lightgaussian: Unbounded 3d gaussian compression with 15x reduction and 200+ fps},
  author={Fan, Zhiwen and Wang, Kevin and Wen, Kairun and Zhu, Zehao and Xu, Dejia and Wang, Zhangyang and others},
  journal={Advances in neural information processing systems},
  volume={37},
  pages={140138--140158},
  year={2024}
}

@inproceedings{chen2024hac,
  title={Hac: Hash-grid assisted context for 3d gaussian splatting compression},
  author={Chen, Yihang and Wu, Qianyi and Lin, Weiyao and Harandi, Mehrtash and Cai, Jianfei},
  booktitle={European Conference on Computer Vision},
  pages={422--438},
  year={2024},
  organization={Springer}
}

@inproceedings{szymanowicz2024splatter,
  title={Splatter image: Ultra-fast single-view 3d reconstruction},
  author={Szymanowicz, Stanislaw and Rupprecht, Chrisitian and Vedaldi, Andrea},
  booktitle={Proceedings of the IEEE/CVF conference on computer vision and pattern recognition},
  pages={10208--10217},
  year={2024}
}

@inproceedings{charatan2024pixelsplat,
  title={pixelsplat: 3d gaussian splats from image pairs for scalable generalizable 3d reconstruction},
  author={Charatan, David and Li, Sizhe Lester and Tagliasacchi, Andrea and Sitzmann, Vincent},
  booktitle={Proceedings of the IEEE/CVF conference on computer vision and pattern recognition},
  pages={19457--19467},
  year={2024}
}

@inproceedings{chen2024mvsplat,
  title={Mvsplat: Efficient 3d gaussian splatting from sparse multi-view images},
  author={Chen, Yuedong and Xu, Haofei and Zheng, Chuanxia and Zhuang, Bohan and Pollefeys, Marc and Geiger, Andreas and Cham, Tat-Jen and Cai, Jianfei},
  booktitle={European Conference on Computer Vision},
  pages={370--386},
  year={2024},
  organization={Springer}
}

@article{xu2024depthsplat,
  title={Depthsplat: Connecting gaussian splatting and depth},
  author={Xu, Haofei and Peng, Songyou and Wang, Fangjinhua and Blum, Hermann and Barath, Daniel and Geiger, Andreas and Pollefeys, Marc},
  journal={arXiv preprint arXiv:2410.13862},
  year={2024}
}

@article{wang2024freesplat,
  title={Freesplat: Generalizable 3d gaussian splatting towards free view synthesis of indoor scenes},
  author={Wang, Yunsong and Huang, Tianxin and Chen, Hanlin and Lee, Gim Hee},
  journal={Advances in Neural Information Processing Systems},
  volume={37},
  pages={107326--107349},
  year={2024}
}

@article{chen2024splatformer,
  title={SplatFormer: Point Transformer for Robust 3D Gaussian Splatting},
  author={Chen, Yutong and Mihajlovic, Marko and Chen, Xiyi and Wang, Yiming and Prokudin, Sergey and Tang, Siyu},
  journal={arXiv preprint arXiv:2411.06390},
  year={2024}
}

@article{ye2024no,
  title={No pose, no problem: Surprisingly simple 3d gaussian splats from sparse unposed images},
  author={Ye, Botao and Liu, Sifei and Xu, Haofei and Li, Xueting and Pollefeys, Marc and Yang, Ming-Hsuan and Peng, Songyou},
  journal={arXiv preprint arXiv:2410.24207},
  year={2024}
}

@inproceedings{ranftl2021vision,
  title={Vision transformers for dense prediction},
  author={Ranftl, Ren{\'e} and Bochkovskiy, Alexey and Koltun, Vladlen},
  booktitle={Proceedings of the IEEE/CVF international conference on computer vision},
  pages={12179--12188},
  year={2021}
}

@article{zhang2022dino,
  title={Dino: Detr with improved denoising anchor boxes for end-to-end object detection},
  author={Zhang, Hao and Li, Feng and Liu, Shilong and Zhang, Lei and Su, Hang and Zhu, Jun and Ni, Lionel M and Shum, Heung-Yeung},
  journal={arXiv preprint arXiv:2203.03605},
  year={2022}
}

@inproceedings{wang2024dust3r,
  title={Dust3r: Geometric 3d vision made easy},
  author={Wang, Shuzhe and Leroy, Vincent and Cabon, Yohann and Chidlovskii, Boris and Revaud, Jerome},
  booktitle={Proceedings of the IEEE/CVF Conference on Computer Vision and Pattern Recognition},
  pages={20697--20709},
  year={2024}
}

@inproceedings{zou2024triplane,
  title={Triplane meets gaussian splatting: Fast and generalizable single-view 3d reconstruction with transformers},
  author={Zou, Zi-Xin and Yu, Zhipeng and Guo, Yuan-Chen and Li, Yangguang and Liang, Ding and Cao, Yan-Pei and Zhang, Song-Hai},
  booktitle={Proceedings of the IEEE/CVF conference on computer vision and pattern recognition},
  pages={10324--10335},
  year={2024}
}

@article{xu2024agg,
  title={Agg: Amortized generative 3d gaussians for single image to 3d},
  author={Xu, Dejia and Yuan, Ye and Mardani, Morteza and Liu, Sifei and Song, Jiaming and Wang, Zhangyang and Vahdat, Arash},
  journal={arXiv preprint arXiv:2401.04099},
  year={2024}
}

@article{tang2024hisplat,
  title={Hisplat: Hierarchical 3d gaussian splatting for generalizable sparse-view reconstruction},
  author={Tang, Shengji and Ye, Weicai and Ye, Peng and Lin, Weihao and Zhou, Yang and Chen, Tao and Ouyang, Wanli},
  journal={arXiv preprint arXiv:2410.06245},
  year={2024}
}

@inproceedings{collins1996space,
  title={A space-sweep approach to true multi-image matching},
  author={Collins, Robert T},
  booktitle={Proceedings CVPR IEEE computer society conference on computer vision and pattern recognition},
  pages={358--363},
  year={1996},
  organization={Ieee}
}

@article{xu2023unifying,
  title={Unifying flow, stereo and depth estimation},
  author={Xu, Haofei and Zhang, Jing and Cai, Jianfei and Rezatofighi, Hamid and Yu, Fisher and Tao, Dacheng and Geiger, Andreas},
  journal={IEEE Transactions on Pattern Analysis and Machine Intelligence},
  volume={45},
  number={11},
  pages={13941--13958},
  year={2023},
  publisher={IEEE}
}

@inproceedings{ronneberger2015u,
  title={U-net: Convolutional networks for biomedical image segmentation},
  author={Ronneberger, Olaf and Fischer, Philipp and Brox, Thomas},
  booktitle={Medical image computing and computer-assisted intervention--MICCAI 2015: 18th international conference, Munich, Germany, October 5-9, 2015, proceedings, part III 18},
  pages={234--241},
  year={2015},
  organization={Springer}
}

@article{yang2024depth,
  title={Depth anything v2},
  author={Yang, Lihe and Kang, Bingyi and Huang, Zilong and Zhao, Zhen and Xu, Xiaogang and Feng, Jiashi and Zhao, Hengshuang},
  journal={Advances in Neural Information Processing Systems},
  volume={37},
  pages={21875--21911},
  year={2024}
}

@article{vaswani2017attention,
  title={Attention is all you need},
  author={Vaswani, Ashish and Shazeer, Noam and Parmar, Niki and Uszkoreit, Jakob and Jones, Llion and Gomez, Aidan N and Kaiser, {\L}ukasz and Polosukhin, Illia},
  journal={Advances in neural information processing systems},
  volume={30},
  year={2017}
}

@inproceedings{liu2021swin,
  title={Swin transformer: Hierarchical vision transformer using shifted windows},
  author={Liu, Ze and Lin, Yutong and Cao, Yue and Hu, Han and Wei, Yixuan and Zhang, Zheng and Lin, Stephen and Guo, Baining},
  booktitle={Proceedings of the IEEE/CVF international conference on computer vision},
  pages={10012--10022},
  year={2021}
}

@article{zhou2018stereo,
  title={Stereo magnification: Learning view synthesis using multiplane images},
  author={Zhou, Tinghui and Tucker, Richard and Flynn, John and Fyffe, Graham and Snavely, Noah},
  journal={arXiv preprint arXiv:1805.09817},
  year={2018}
}

@inproceedings{liu2021infinite,
  title={Infinite nature: Perpetual view generation of natural scenes from a single image},
  author={Liu, Andrew and Tucker, Richard and Jampani, Varun and Makadia, Ameesh and Snavely, Noah and Kanazawa, Angjoo},
  booktitle={Proceedings of the IEEE/CVF International Conference on Computer Vision},
  pages={14458--14467},
  year={2021}
}

@inproceedings{jensen2014large,
  title={Large scale multi-view stereopsis evaluation},
  author={Jensen, Rasmus and Dahl, Anders and Vogiatzis, George and Tola, Engin and Aan{\ae}s, Henrik},
  booktitle={Proceedings of the IEEE conference on computer vision and pattern recognition},
  pages={406--413},
  year={2014}
}

@inproceedings{dai2017scannet,
  title={Scannet: Richly-annotated 3d reconstructions of indoor scenes},
  author={Dai, Angela and Chang, Angel X and Savva, Manolis and Halber, Maciej and Funkhouser, Thomas and Nie{\ss}ner, Matthias},
  booktitle={Proceedings of the IEEE conference on computer vision and pattern recognition},
  pages={5828--5839},
  year={2017}
}

@inproceedings{ling2024dl3dv,
  title={Dl3dv-10k: A large-scale scene dataset for deep learning-based 3d vision},
  author={Ling, Lu and Sheng, Yichen and Tu, Zhi and Zhao, Wentian and Xin, Cheng and Wan, Kun and Yu, Lantao and Guo, Qianyu and Yu, Zixun and Lu, Yawen and others},
  booktitle={Proceedings of the IEEE/CVF Conference on Computer Vision and Pattern Recognition},
  pages={22160--22169},
  year={2024}
}

@inproceedings{zhang2025transplat,
  title={Transplat: Generalizable 3d gaussian splatting from sparse multi-view images with transformers},
  author={Zhang, Chuanrui and Zou, Yingshuang and Li, Zhuoling and Yi, Minmin and Wang, Haoqian},
  booktitle={Proceedings of the AAAI Conference on Artificial Intelligence},
  volume={39},
  number={9},
  pages={9869--9877},
  year={2025}
}

@article{loshchilov2017decoupled,
  title={Decoupled weight decay regularization},
  author={Loshchilov, Ilya and Hutter, Frank},
  journal={arXiv preprint arXiv:1711.05101},
  year={2017}
}

@inproceedings{gu2020cascade,
  title={Cascade cost volume for high-resolution multi-view stereo and stereo matching},
  author={Gu, Xiaodong and Fan, Zhiwen and Zhu, Siyu and Dai, Zuozhuo and Tan, Feitong and Tan, Ping},
  booktitle={Proceedings of the IEEE/CVF conference on computer vision and pattern recognition},
  pages={2495--2504},
  year={2020}
}

@article{he2026surfsplat,
  title={SurfSplat: Conquering Feedforward 2D Gaussian Splatting with Surface Continuity Priors},
  author={He, Bing and Gao, Jingnan and Chen, Yunuo and Cao, Ning and Chen, Gang and Cheng, Zhengxue and Song, Li and Zhang, Wenjun},
  journal={arXiv preprint arXiv:2602.02000},
  year={2026}
}

@inproceedings{kang2025selfsplat,
  title={Selfsplat: Pose-free and 3d prior-free generalizable 3d gaussian splatting},
  author={Kang, Gyeongjin and Yoo, Jisang and Park, Jihyeon and Nam, Seungtae and Im, Hyeonsoo and Shin, Sangheon and Kim, Sangpil and Park, Eunbyung},
  booktitle={Proceedings of the Computer Vision and Pattern Recognition Conference},
  pages={22012--22022},
  year={2025}
}

@article{jiang2025anysplat,
  title={Anysplat: Feed-forward 3d gaussian splatting from unconstrained views},
  author={Jiang, Lihan and Mao, Yucheng and Xu, Linning and Lu, Tao and Ren, Kerui and Jin, Yichen and Xu, Xudong and Yu, Mulin and Pang, Jiangmiao and Zhao, Feng and others},
  journal={ACM Transactions on Graphics (TOG)},
  volume={44},
  number={6},
  pages={1--16},
  year={2025},
  publisher={ACM New York, NY, USA}
}

@misc{ye2025yonosplatneedmodelfeedforward,
      title={YoNoSplat: You Only Need One Model for Feedforward 3D Gaussian Splatting}, 
      author={Botao Ye and Boqi Chen and Haofei Xu and Daniel Barath and Marc Pollefeys},
      year={2025},
      eprint={2511.07321},
      archivePrefix={arXiv},
      primaryClass={cs.CV},
      url={https://arxiv.org/abs/2511.07321}, 
}

@article{dens3r,
      title={Dens3R: A Foundation Model for 3D Geometry Prediction}, 
      author={Xianze Fang and Jingnan Gao and Zhe Wang and Zhuo Chen and Xingyu Ren and Jiangjing Lyu and Qiaomu Ren and Zhonglei Yang and Xiaokang Yang and Yichao Yan and Chengfei Lyu},
      journal={arXiv preprint arXiv:2507.16290},
      year={2025}
}

@article{more,
  title={MoRE: 3D Visual Geometry Reconstruction Meets Mixture-of-Experts}, 
  author={Jingnan Gao and Zhe Wang and Xianze Fang and Xingyu Ren and Zhuo Chen and Shengqi Liu and Yuhao Cheng and Jiangjing Lyu and Xiaokang Yang and Yichao Yan},
  journal={arXiv preprint arXiv:2510.27234},
  year={2025}
}

@inproceedings{wang2025vggt,
  title={Vggt: Visual geometry grounded transformer},
  author={Wang, Jianyuan and Chen, Minghao and Karaev, Nikita and Vedaldi, Andrea and Rupprecht, Christian and Novotny, David},
  booktitle={2025 IEEE/CVF Conference on Computer Vision and Pattern Recognition (CVPR)},
  pages={5294--5306},
  year={2025},
  organization={IEEE}
}

@article{ren2026tokengs,
  title={Tokengs: Decoupling 3d gaussian prediction from pixels with learnable tokens},
  author={Ren, Jiawei and Tyszkiewicz, Michal Jan and Huang, Jiahui and Gojcic, Zan},
  journal={arXiv preprint arXiv:2604.15239},
  year={2026}
}

@article{an2025c3g,
  title={C3g: Learning compact 3d representations with 2k gaussians},
  author={An, Honggyu and Jung, Jaewoo and Kim, Mungyeom and Kim, Chaehyun and Jeon, Minkyeong and Han, Jisang and Fukuda, Kazumi and Narihira, Takuya and Ko, Hyuna and Kim, Junsu and others},
  journal={arXiv preprint arXiv:2512.04021},
  year={2025}
}

@inproceedings{wang2026volsplat,
  title={Volsplat: Rethinking feed-forward 3d gaussian splatting with voxel-aligned prediction},
  author={Wang, Weijie and Chen, Yeqing and Zhang, Zeyu and Liu, Hengyu and Wang, Haoxiao and Feng, Zhiyuan and Qin, Wenkang and Chen, Feng and Bian, Jia-Wang and Zhu, Zheng and others},
  booktitle={European Conference on Computer Vision},
  pages={1--19},
  year={2026},
  organization={Springer}
}

@inproceedings{moreau2026off,
  title={Off the grid: Detection of primitives for feed-forward 3d gaussian splatting},
  author={Moreau, Arthur and Shaw, Richard and Nazarczuk, Michal and Shin, Jisu and Tanay, Thomas and Zhang, Zhensong and Xu, Songcen and P{\'e}rez-Pellitero, Eduardo},
  booktitle={Proceedings of the IEEE/CVF Conference on Computer Vision and Pattern Recognition},
  pages={11756--11766},
  year={2026}
}

@inproceedings{tang2025mv,
  title={Mv-dust3r+: Single-stage scene reconstruction from sparse views in 2 seconds},
  author={Tang, Zhenggang and Fan, Yuchen and Wang, Dilin and Xu, Hongyu and Ranjan, Rakesh and Schwing, Alexander and Yan, Zhicheng},
  booktitle={2025 IEEE/CVF Conference on Computer Vision and Pattern Recognition (CVPR)},
  pages={5283--5293},
  year={2025},
  organization={IEEE}
}
}
\newpage
\appendix

\section{Technical Appendices and Supplementary Material}

\vspace{-5pt}
\subsection{Preliminaries}

Generalizable 3D Gaussian Splatting (3DGS) frameworks bypass the computationally heavy per-scene optimization by directly inferring a 3D scene representation from a sparse set of observations. Operating in a single forward pass, these networks typically adopt a dense, pixel-aligned prediction paradigm. Specifically, given $V$ source views $\{I^v\}_{v=1}^{V}$ along with their corresponding camera intrinsics $\{\mathbf{k}^v\}_{v=1}^{V}$ and extrinsics $\{\mathbf{T}^v\}_{v=1}^{V}$, a standard feedforward model $f_\theta$ outputs a set of Gaussian primitives for every pixel coordinate:

\begin{equation}
    f_\theta : \{(I^v, \mathbf{k}^v, \mathbf{T}^v)\}_{v=1}^{V} \mapsto \left\{ \bigcup_{j=1}^{H \times W} \left(\boldsymbol{\mu}_j^v, \boldsymbol{\alpha}_j^v, \mathbf{r}_j^v, \mathbf{s}_j^v, \mathbf{c}_j^v\right) \right\}_{v=1}^{V},
\end{equation}

where $H$ and $W$ represent the spatial dimensions of the input images. The tuple defines the explicit attributes of the $j$-th Gaussian originating from view $v$: its 3D center coordinate $\boldsymbol{\mu}_j^v$, opacity $\boldsymbol{\alpha}_j^v$, rotation quaternion $\mathbf{r}_j^v$, scaling factor $\mathbf{s}_j^v$, and spherical harmonics $\mathbf{c}_j^v$.

The feasibility of this dense prediction relies heavily on the robust geometric priors extracted by modern vision backbones (e.g., ViTs~\cite{ranftl2021vision,zhang2022dino,wang2024dust3r,dens3r,more,wang2025vggt}). By unprojecting these rich 2D features into 3D space using camera parameters, the network establishes a rigid pixel-to-Gaussian mapping that can be supervised end-to-end via differentiable rasterization. 

However, this rigid $H \times W$ formulation indiscriminately populates the 3D space with primitives regardless of the underlying scene complexity. By forcing a strict 1:1 mapping between pixels and Gaussians, existing architectures inevitably generate severe spatial redundancy in homogeneous regions and fail to exploit the geometric priors of piecewise planar surfaces. It is precisely this fundamental bottleneck in representational efficiency that CompactSplat seeks to overcome.

\subsection{Encoder Architecture}
\label{encoder}
We adopt a dual-branch encoder design to extract both monocular and multi-view features for robust 3D reasoning, following the architecture proposed by DepthSplat~\cite{xu2024depthsplat} and subsequent SurfSPlat~\cite{he2026surfsplat}.

\paragraph{Multi-view Branch.}
The multi-view encoder begins with a lightweight ResNet-style backbone composed of stride-2 convolutional layers, yielding spatially downsampled feature maps by a factor of \( s \). To enable view aggregation, we employ a multi-view Swin Transformer~\cite{liu2021swin} consisting of 6 stacked self- and cross-attention layers. This module outputs multi-view-aware features \(\left\{\boldsymbol{F}^i\right\}_{i=1}^N\), where \(\boldsymbol{F}^i \in \mathbb{R}^{\frac{H}{s} \times \frac{W}{s} \times C}\).

We further adopt the plane-sweep stereo technique~\cite{collins1996space,xu2023unifying} to construct geometric consistency. We uniformly sample \( D \) candidate depths between near and far bounds. Given reference view \( i \) and source view \( j \), we warp features \(\boldsymbol{F}_j\) to view \( i \) at each depth \( d_m \), resulting in \(\left\{\boldsymbol{F}_{d_m}^{j \rightarrow i}\right\}_{m=1}^D\). These warped volumes are compared to \(\boldsymbol{F}_i\) via dot-product similarity to construct a cost volume \(\boldsymbol{C}^i \in \mathbb{R}^{\frac{H}{s} \times \frac{W}{s} \times D}\).

\paragraph{Single-view Branch.}
We utilize the ViT backbone from Depth Anything V2 model~\cite{yang2024depth} to extract monocular features. The output has a spatial resolution of \(1/14\) relative to the original image and is bilinearly upsampled to match the cost volume resolution, yielding monocular features \(\boldsymbol{F}_{\text{m}}^i \in \mathbb{R}^{\frac{H}{s} \times \frac{W}{s} \times C_\text{m}}\).

\paragraph{U-Net and Depth Prediction.}
The monocular and multi-view features \(\boldsymbol{F}_\text{m}^i\) and \(\boldsymbol{C}^i\) are concatenated along the channel dimension and processed by a 2D U-Net to produce depth candidates \(\boldsymbol{D}^i \in \mathbb{R}^{\frac{H}{s} \times \frac{W}{s} \times D}\). A softmax operation is applied over the depth axis, followed by a weighted summation to generate the predicted depth map.

To enhance depth quality, we employ a hierarchical cascade structure~\cite{gu2020cascade}, refining the predicted depth to \(\boldsymbol{D}_{ds}^i \in \mathbb{R}^{\frac{2H}{s} \times \frac{2W}{s}}\), which is subsequently upsampled to full resolution using a DPT head~\cite{ranftl2021vision}.

\paragraph{Attribute Prediction.}
The predicted depth is used to reconstruct Gaussian positions. For estimating the remaining Gaussian attributes—such as scale multipliers, high-frequency SH coefficients, and opacity—we apply an additional DPT head, conditioned on a concatenation of the input image, predicted depth, and encoder features.

\paragraph{Hyperparameter Selection.}
The downsample scale $s$ is set to 4. Channel number $C$ is set to 128, channel number $D$ is set to 128. The channel number $C_\text{m}$ of the monocular feature is set to 64 for small model, 96 for base model, 128 for large model.

\paragraph{Gaussian Processor.}The Surface Continuity Prior(SCP) and the Forced Alpha Blending methods from Surfsplat have been adopted to process the gaussian attributes.

\vspace{0.5em}
\noindent\textbf{Note:} Our implementation is consistent with DepthSplat~\cite{xu2024depthsplat} and SurfSplat~\cite{he2026surfsplat} for reproducibility. No architectural modifications are made to the encoder unless otherwise stated.

\refstepcounter{mypara} \label{para:reproducibility}
\subsection{Implementation Details.}
CompactSplat is implemented in PyTorch~\cite{paszke2019pytorch} and optimized using the AdamW optimizer~\cite{loshchilov2017decoupled} coupled with a cosine annealing learning rate schedule. For the core components of our proposed architecture, we set the maximum block size in the Texture-Aware Partitioner to $S_{\text{max}} = 16$. During the sparsity-aware fine-tuning stage of our curriculum, the dynamic partition threshold is sampled uniformly up to $\tau_{\text{max}} = 0.02$.

Following standard practices~\cite{he2026surfsplat,xu2024depthsplat}, we instantiate our model using varying capacities of the Depth Anything V2~\cite{yang2024depth} monocular backbone (ViT-S, ViT-B, and ViT-L), denoting these variants as Ours-S, Ours-B, and Ours-L, respectively. All models are trained on NVIDIA A100 GPUs. Specifically, Ours-S is trained for 300K iterations with a batch size of 16, whereas Ours-B and Ours-L are trained for 600K iterations with a batch size of 8. The learning rate is configured at $2 \times 10^{-6}$ for the pretrained monocular backbone and $2 \times 10^{-4}$ for all newly initialized layers.

\subsection{Cross-Dataset Generalization}
As detailed in Table~\ref{cross}, our extended cross-dataset evaluations underscore the robust generalizability of CompactSplat. Even when evaluated outside its primary training distribution, CompactSplat achieves rendering fidelity comparable to the state-of-the-art SurfSplat~\cite{he2026surfsplat} baseline, while consistently yielding a significantly more compact and memory-efficient 3D representation.

\begin{table}[h]
\centering
\caption{Cross datasets performance.}
\label{cross}
\fontsize{5}{9.5}\selectfont
\setlength{\tabcolsep}{3.5pt}
\begin{tabular}{lcccc|cccc|cccc|cccc}
\toprule
& \multicolumn{4}{c|}{Scannet} & \multicolumn{4}{c|}{DL3DV} & \multicolumn{4}{c|}{DTU} & \multicolumn{4}{c}{Average} \\
Method & PSNR↑ & SSIM↑ & LPIPS↓ & GPP↓ & PSNR↑ & SSIM↑ & LPIPS↓ & GPP↓ & PSNR↑ & SSIM↑ & LPIPS↓ & GPP↓ & PSNR↑ & SSIM↑ & LPIPS↓ & GPP↓ \\
\midrule
pixelSplat~\citeyear{charatan2024pixelsplat}  & 19.606 & 0.714 & 0.324 & 3.000 & 27.201 & 0.882 & 0.104 & 3.000 & 12.752 & 0.329 & 0.639 & 3.000 & 19.853 & 0.642 & 0.356 & 3.000 \\
HiSplat~\citeyear{tang2024hisplat}  & 19.095 & 0.691 & 0.342 & 1.750 & 26.242 & 0.869 & 0.112 & 1.750 & 16.019 & 0.671 & 0.277 & 1.750 & 20.452 & 0.744 & 0.244 & 1.750 \\
\midrule
MVSplat~\citeyear{chen2024mvsplat} & 18.725 & 0.692 & 0.333 & 1.000 & 23.841 & 0.768 & 0.156 & 1.000 & 13.914 & 0.470 & 0.386 & 1.000 & 18.827 & 0.643 & 0.292 & 1.000 \\
TranSplat~\citeyear{zhang2025transplat}  & 18.944 & 0.705 & 0.332 & 1.000 & 23.913 & 0.771 & 0.161 & 1.000 & 14.956 & 0.527 & 0.327 & 1.000 & 19.271 & 0.668 & 0.273 & 1.000  \\
DepthSplat~\citeyear{xu2024depthsplat}  & 20.201 & 0.735 & 0.305 & 1.000 & 28.141 & 0.905 & 0.083 & 1.000 & 14.592 & 0.425 & 0.436 & 1.000 & 20.978 & 0.688 & 0.275 & 1.000 \\
SurfSplat~\citeyear{he2026surfsplat} & 20.305 & 0.731 & 0.313 & 1.000 & 27.384 & 0.890 & 0.106 & 1.000 & 15.544 & 0.488 & 0.329 & 1.000 & 21.078 & 0.703 & 0.249 & 1.000 \\
YonoSplat$^*$~\citeyear{ye2025yonosplatneedmodelfeedforward} & 20.323 & 0.727 & 0.311 & 1.000 & 25.636 & 0.843 & 0.124 & 1.000 & 11.721 & 0.343 & 0.596 & 1.000 & 19.227 & 0.638 & 0.261 & 1.000 \\
Ours & 19.905 & 0.706 & 0.353 & 0.643 & 25.573 & 0.868 & 0.152 & 0.715 & 15.368 & 0.476 & 0.348 & 0.698 & 20.208 & 0.683 & 0.284 & 0.685 \\
\bottomrule
\end{tabular}
\vspace{-10pt}
\end{table}

\subsection{Comparison with other non-pixel-aligned methods}
\begin{table}[ht]
\centering
\caption{Comparison with non-pixel-aligned methods on RE10K.} 
\label{tab:nonpixel}
\begin{tabular}{lrrrr}
\toprule
Method & PSNR$\uparrow$ & SSIM$\uparrow$ & LPIPS$\downarrow$ & GS Num.$\downarrow$ \\
\midrule
TokenGS~\cite{ren2026tokengs} & 28.02 & 0.896 & 0.147 & 66k \\
C3G~\cite{an2025c3g} & 22.39 & 0.713 & 0.259 & 2k \\
VolSplat~\cite{wang2026volsplat} & 22.86 & 0.811 & 0.201 & 131k \\
CompactSplat-L & 27.57 & 0.892 & 0.117 & 72.6k \\
\bottomrule
\end{tabular}
\end{table}

TokenGS (1024 tokens) and C3G use published results; VolSplat uses our two-view re-evaluation. TokenGS and CompactSplat use two $256 \mathbin{\times}256$ inputs. C3G reports $224 \mathbin{\times}224$ training inputs and uses test-time target-camera alignment. Training and camera handling differ across methods.

\subsection{Extended Qualitative Results}
\begin{figure}[ht]
  \vspace{-12pt}
  \centering
  \includegraphics[width=1\linewidth]{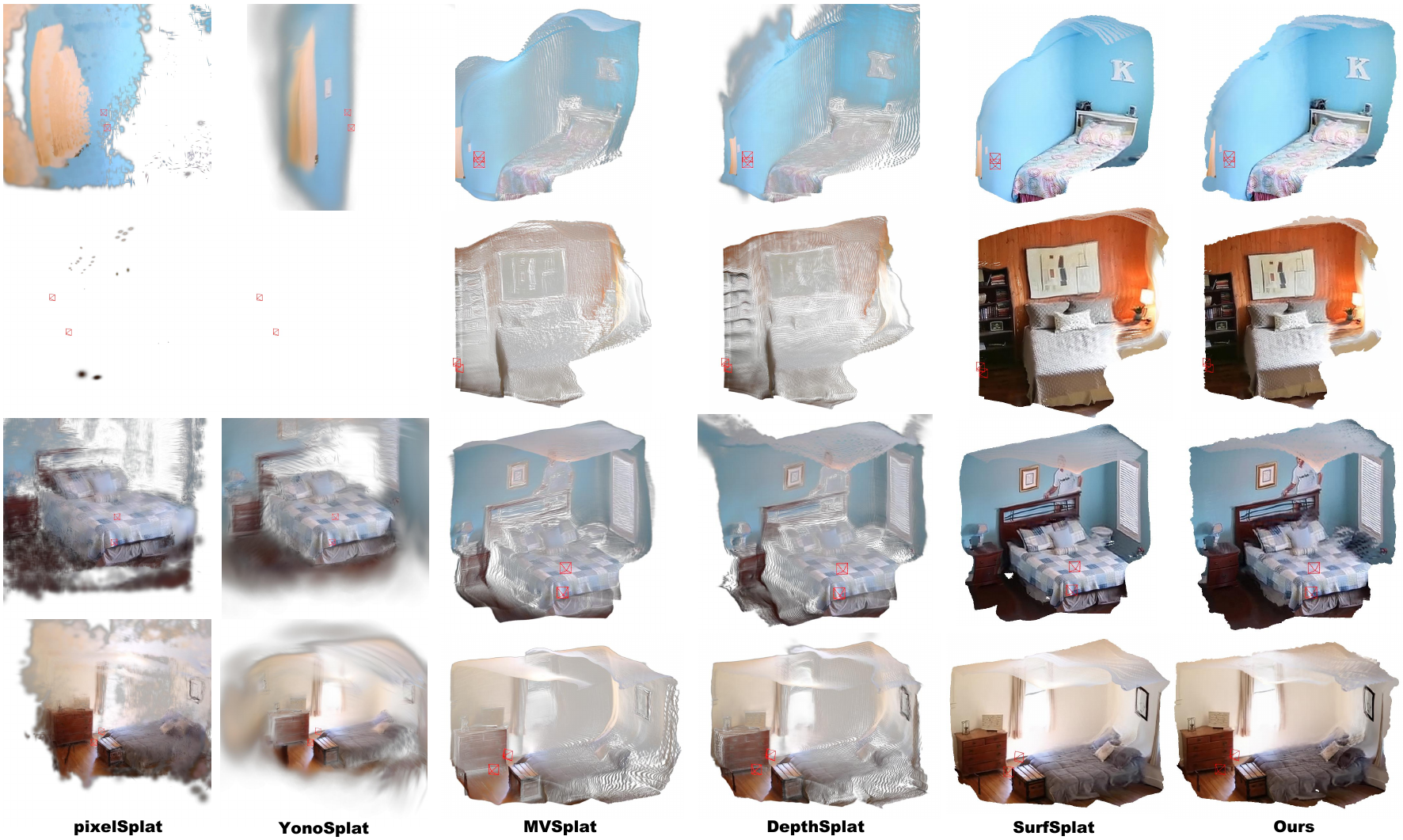}
  \caption{\textbf{More subjective comparison of generated 3D scenes.} 
    }
  \label{fig:app}
  \vspace{-10pt}
\end{figure}

Figure~\ref{fig:app} presents an additional set of reconstructed scenes across diverse environments. These qualitative visualizations further corroborate that CompactSplat reliably maintains photorealistic rendering quality and watertight geometric structures, all while empowering users with dynamic, on-the-fly control over representational compactness.

\refstepcounter{mypara} \label{para:limitation}
\subsection{Limitations and Future Directions.}

While CompactSplat significantly improves representational efficiency, it leaves exciting room for future exploration. First, although our quadtree-based partitioning is highly efficient, it relies on rigid orthogonal grids. Exploring more sophisticated, shape-aware, or semantic-driven subdivision strategies  could further push the boundaries of the Pareto frontier, extracting even higher visual fidelity at extreme pruning ratios. Second, our current formulation primarily eliminates spatial redundancy within individual views, leaving the inherent \textit{inter-view} redundancy largely unexplored. Since multiple source cameras inevitably observe overlapping geometries, future research could investigate cross-view primitive pruning or multi-view fusion mechanisms to further condense the global 3D scene representation.
\end{document}